\documentclass[final,5p,times,twocolumn]{elsarticle}

\usepackage{hyperref}
\usepackage{subfig}
\usepackage{multirow}
\usepackage[fleqn]{amsmath}
\usepackage{booktabs}
\usepackage{algorithmic}
\usepackage{enumerate}
\usepackage{array}
\usepackage{gensymb}
\usepackage{stfloats}
\fnbelowfloat
\usepackage{url}
\usepackage{color}
\usepackage{stfloats}
\usepackage{amssymb}
\usepackage{arydshln}

\usepackage{booktabs}
\usepackage{array, caption, threeparttable}
\usepackage[font=small,labelfont=bf,labelsep=none]{caption}
\usepackage[marginal]{footmisc}

\usepackage{enumitem}
\usepackage{siunitx}

\usepackage{lineno}

\usepackage{listings}
\usepackage{xcolor}
\usepackage{float}

\usepackage[ruled,vlined]{algorithm2e}

\begin{document}

	\begin{frontmatter}

        \title{SVII-3D: Advancing Roadside Infrastructure Inventory with Decimeter-level \\3D Localization and Comprehension from Sparse Street Imagery}
        
		\author[liesmars]{Chong Liu}

        \author[liesmars]{Luxuan Fu}

        \author[sichuan]{Yang Jia}

        \author[liesmars]{Zhen Dong\corref{correspondingauthor}}
        
        \author[liesmars]{Bisheng Yang}
		
		\cortext[correspondingauthor]{Corresponding authors.}
		\address[liesmars]{State Key Laboratory of Information Engineering in Surveying, Mapping and Remote Sensing (LIESMARS), Wuhan University, Wuhan 430079, China}
        \address[sichuan]{Sichuan Highway Planning, Survey, Design and Research Institute Ltd, Chengdu 610000, China}
	
		\begin{abstract}
            The automated creation of digital twins and precise asset inventories is a critical task in smart city construction and facility lifecycle management. However, utilizing cost-effective sparse imagery remains challenging due to limited robustness, inaccurate localization, and a lack of fine-grained state understanding. To address these limitations, SVII-3D, a unified framework for holistic asset digitization, is proposed. First, LoRA fine-tuned open-set detection is fused with a spatial-attention matching network to robustly associate observations across sparse views. Second, a geometry-guided refinement mechanism is introduced to resolve structural errors, achieving precise decimeter-level 3D localization. Third, transcending static geometric mapping, a Vision-Language Model agent leveraging multi-modal prompting is incorporated to automatically diagnose fine-grained operational states. Experiments demonstrate that SVII-3D significantly improves identification accuracy and minimizes localization errors. Consequently, this framework offers a scalable, cost-effective solution for high-fidelity infrastructure digitization, effectively bridging the gap between sparse perception and automated intelligent maintenance.
		\end{abstract}


	\end{frontmatter}
	
	\section{Introduction}
            In the lifecycle of civil infrastructure, the Operations and Maintenance (O\&M) phase requires accurate, up-to-date 3D information to support decision-making. To advance digital construction and intelligent management, there is an urgent need for high-precision 3D perception and understanding \citep{dong2025neural} under low-cost acquisition conditions. This requires creating comprehensive databases that accurately capture the 3D spatial locations, semantic categories, and fine-grained attributes and states of infrastructure assets \citep{liu2025training}.
            However, current solutions face a dilemma: LiDAR-based methods offer high geometric precision \citep{justo2021scan} but incur prohibitive hardware and operational costs, while existing image-based approaches are cost-effective yet typically suffer from meter-level localization errors.
            To resolve this trade-off, this study investigates how to leverage sparse, low-cost street-view imagery and vision-language models (VLMs) to enable automated, high-precision 3D localization of multi-class infrastructure targets, and further to extract their key attributes and operational states, providing deeper semantic support for subsequent management and decision-making.

            Most existing infrastructure perception approaches rely on Google Street View imagery or crowd-sourced street-level images, employing object detection, matching, and depth estimation to obtain 2D latitude–longitude coordinates. While these methods have facilitated large-scale mapping, they generally suffer from several limitations.
            First, they struggle with robust infrastructure identification. Not only are they often category-specific and data-hungry \citep{wang2024few}, but their reliance on local frame-to-frame matching also lacks global reasoning, leading to error propagation and inaccurate asset counting (e.g., fragmentation or erroneous merges) \citep{krylov2018automatic}.
            Third, their output is generally restricted to 2D coordinates with meter-level errors \citep{wilson2022object}, lacking precise 3D localization.
            Finally, they overlook fine-grained semantic attributes, failing to provide the operational status context required for intelligent management.
            
            Although advanced paradigms like dense 3D reconstruction and multi-object tracking excel in video processing, their direct application to street-view imagery is hindered by a fundamental constraint: data sparsity. Discrete capture intervals create wide baselines, rendering these methods ineffective. specifically, dense reconstruction methods (e.g., MVS \citep{schonberger2016pixelwise}, NeRF \citep{mildenhall2021nerf}, 3D Gaussian Splatting \citep{kerbl20233d} and VGGT \citep{wang2025vggt}) fail to synthesize valid geometry due to insufficient overlap, while multi-object tracking (MOT) algorithms (e.g., DeepSORT \citep{wojke2017simple}, TrackFormer \citep{meinhardt2022trackformer} and MambaTrack \citep{xiao2024mambatrack}) collapse as drastic scale and appearance changes violate their continuity assumptions. Consequently, robust cross-view association in such sparse scenarios remains an open challenge.

            To address the above challenges, this study proposes SVII-3D (\textbf{S}parse \textbf{V}isual \textbf{I}nfrastructure \textbf{I}nventory in \textbf{3D}), a novel framework for infrastructure 3D localization and understanding from sparse street-view imagery. First, we adopt a large open-set object detection model and apply LoRA fine-tuning with a small amount of labeled data. Building upon this, we design a spatial-attention-enhanced Transformer matching network to establish robust associations across multi-view observations. Furthermore, we introduce a geometry-guided 3D localization strategy and a matching refinement mechanism. Finally, we extend the framework with a state-discriminative VLM agent, which interprets the fine-grained attributes or operational states of infrastructure assets.
            The main contributions of this study are summarized as follows:
            \begin{itemize}
            \item \textbf{Robust Infrastructure Identification}: We propose a unified identification pipeline that synergizes LoRA fine-tuned open-set detection with a spatial-attention matching network. This combination enables accurate discovery and robust cross-view grouping of diverse infrastructure assets, even under limited annotation and sparse imagery conditions.
            \item \textbf{Geometry-guided 3D Localization with Global Refinement}: We design a geometry-guided 3D localization strategy combined with a matching refinement mechanism, which jointly leverages geometric priors and spatial reasoning to solve over-matching and under-matching, ensuring precise 3D localization of infrastructure assets.
            \item \textbf{State-discriminative VLM Agent}: We develop a state-discriminative vision–language agent that incorporates task-specific multi-modal prompt engineering, retrieval augmentation and expert knowledge injection, enabling the model to infer fine-grained states and semantic attributes of infrastructure targets. This enriches the asset inventory by providing operational status (e.g., structural health), directly supporting automated maintenance inspections.
            \end{itemize}
            
        \section{Related work}
            
            Research on infrastructure 3D localization from multi-view or street-view imagery generally follows a three-step pipeline: detection, matching, and localization \citep{wilson2024image}. 
            First, infrastructure objects are identified in single images using detection or segmentation models.
            Second, these observations must be reliably matched across multiple views.
            Finally, the matched observations are anchored to the real-world coordinate system through geo-localization.
            However, existing methods typically focus only on category-level identification of infrastructure objects, without extracting richer attributes and operational states, thus lacking the contextual knowledge necessary for comprehensive asset inventory.
            
            \subsection{Image infrastructure detection}
                The prerequisite for infrastructure 3D localization is accurate 2D detection from street-level imagery. Over the past decade, this field has been propelled by large-scale benchmarks. General-purpose datasets like Cityscapes \citep{cordts2016cityscapes} and Objects365 \citep{shao2019objects365} provide extensive annotations for pixel-level scene understanding and generic object detection, covering common classes like street lights and cones. Simultaneously, domain-specific resources such as GTSDB \citep{houben2013detection} and TT100K \citep{zhu2016traffic} offer fine-grained attributes for traffic signs. Building on these foundations, numerous studies have developed advanced detectors to automate infrastructure extraction \citep{cui2020context, jain2023oneformer, erisen2024sernet}.

                However, these approaches operate under a closed-set assumption, restricting detection to predefined categories and necessitating costly re-annotation for new infrastructure types. To address this rigidity, open-vocabulary detection (OVD) has emerged as a flexible alternative \citep{li2022grounded, minderer2022simple, cheng2024yolo}. By aligning visual features with arbitrary textual \citep{ren2024grounding} or visual prompts \citep{jiang2024t}, models like Grounding DINO \citep{ren2024grounding} and DINO-X \citep{ren2024dino} demonstrate remarkable category-agnostic capabilities.
                Nevertheless, these foundation models are primarily trained on general-purpose data. Consequently, their zero-shot performance often degrades on domain-specific infrastructure (e.g., surveillance cameras, bollards) due to the significant domain gap, necessitating efficient adaptation strategies.

            \subsection{Target observations matching}
                After obtaining 2D observations of infrastructure in individual images, the second step involves aggregating and matching these observations across multiple views. 
                Existing approaches can be broadly categorized into three families:
                
                \textbf{Geometry-based projection and consistency methods.}  
                Early works \citep{timofte2011multi, hebbalaguppe2017telecom} rely on geometric consistency across views. 
                \citep{krylov2018automatic} represent detections as 3D rays and employ an MRF formulation that integrates spatial and depth constraints to disambiguate and validate the intersections corresponding to actual objects. 
                \citep{nassar2019simultaneous} use geo-metadata to project detections across views, refine the projections with a learned network, and associate observations based on overlapping boxes.
                However, these methods typically rely on an object sparsity assumption that instances are separated by at least a minimum distance which is often violated in dense urban corridors, leading to ambiguous associations and false merges.
                
                \textbf{Graph-based reasoning methods.}  
                Instead of pairwise matching, \citep{nassar2020geograph} represent detections as nodes enriched with both appearance features extracted from the detector and geometric attributes derived from camera geo-metadata, while edges encode cross-view consistency cues. A graph neural network is then trained to jointly reason over this structure and aggregate observations into consistent object hypotheses. However, such graph-based reasoning can be highly brittle: a single spurious edge between two detections may erroneously merge them into the same object hypothesis.
                
                \textbf{Similarity learning methods.}
                Recent approaches directly predict cross-view affinity by fusing visual embeddings with geometric cues (e.g., GPS metadata) \citep{chaabane2021end, wilson2022object}. These methods typically formulate correspondence as a bipartite matching problem between adjacent frames. However, they fundamentally limit reasoning to local pairwise interactions, lacking a global view of the multi-frame graph. Consequently, long-range associations must be inferred through sequential chaining—a process inherently prone to error accumulation, where a single missed link or mismatch causes the entire trajectory to fracture or drift.

            \begin{figure*}[!h]
            \centering{
            \includegraphics[width=0.95\textwidth]{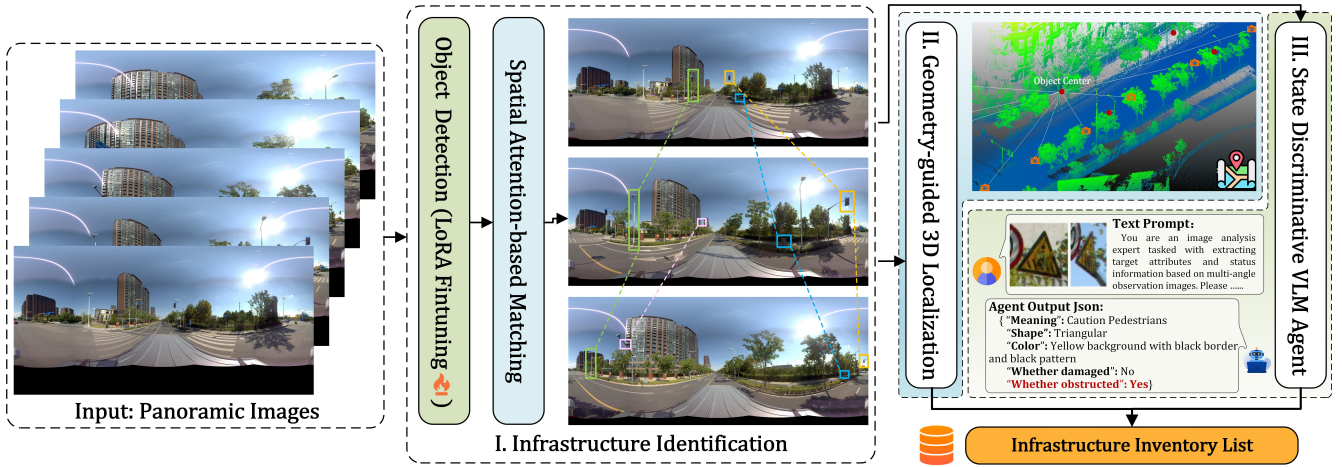}}%
            \caption {Overview of the proposed SVII-3D framework. The system operates in three stages: (I) Infrastructure Identification, combining LoRA fine-tuned detection and spatial-attention matching to associate sparse observations; (II) Geometry-guided 3D Localization, which refines associations for decimeter-level positioning; and (III) State-discriminative VLM Agent, inferring semantic attributes and operational states. This pipeline generates a precise, semantic-rich infrastructure inventory list.}
            \label {fig_workflow}
            \end{figure*}
                
            \subsection{Infrastructure geo-localization}
                The final step anchors the matched observations to the real-world coordinate system through geo-localization, leveraging GPS/IMU metadata, camera poses, or geometric triangulation to estimate accurate 3D positions. Existing approaches can be broadly categorized into two types.

                The first type relies on multi-view geometric methods, which explicitly utilize observations from multiple viewpoints \citep{krylov2018object, kim2019participatory}. By triangulating corresponding detections across different frames or sensors, they directly recover 3D positions that are consistent with the physical environment \citep{hebbalaguppe2017telecom,krylov2018automatic,nassar2019simultaneous}. These methods leverage strong geometric constraints, but typically require accurate calibration and sufficient viewpoint diversity.

                The second type comprises methods that estimate 3D positions from single-view observations \citep{campbell2019detecting}, relying either on learning-based regression or depth estimation \citep{yang2024depth, yang2024depthv2} approaches, with visual cues and auxiliary metadata as input. The estimated locations of observations belonging to the same object are then fused using simple heuristics such as averaging, without jointly reasoning across multiple views \citep{nassar2020geograph,chaabane2021end,wilson2022object}. While this strategy avoids strict geometric prerequisites, it underutilizes the complementary information across multiple views, thereby leading to reduced localization accuracy. In addition, these methods often sacrifice interpretability, depend heavily on the training data distribution, and consequently exhibit limited generalization capability.
        
        \section{Methodology}
                As illustrated in Fig.~\ref{fig_workflow}, the proposed \textbf{SVII-3D} framework establishes a unified pipeline for roadside infrastructure inventory from sparse street-level imagery. 
                The process begins with \textbf{Infrastructure Identification}, where LoRA fine-tuned open-set detection is synergized with a spatial-attention Transformer to robustly discover and associate unique object instances across wide-baseline views.
                Building on these associations, we employ \textbf{Geometry-guided 3D Localization}, which utilizes multi-view ray triangulation and a global refinement mechanism to correct structural matching errors, achieving accurate counting and decimeter-level positioning.
                Finally, the system advances to semantic understanding via a \textbf{State-discriminative VLM Agent}, inferring fine-grained operational states through multi-modal prompting, retrieval augmentation, and expert knowledge injection. 
                The resulting inventory delivers precise 3D locations and maintenance-ready semantic judgments, directly supporting automated inspection and decision-making in real-world infrastructure management.

            \begin{figure*}[!h]
                \centering{
                \includegraphics[width=0.95\textwidth]{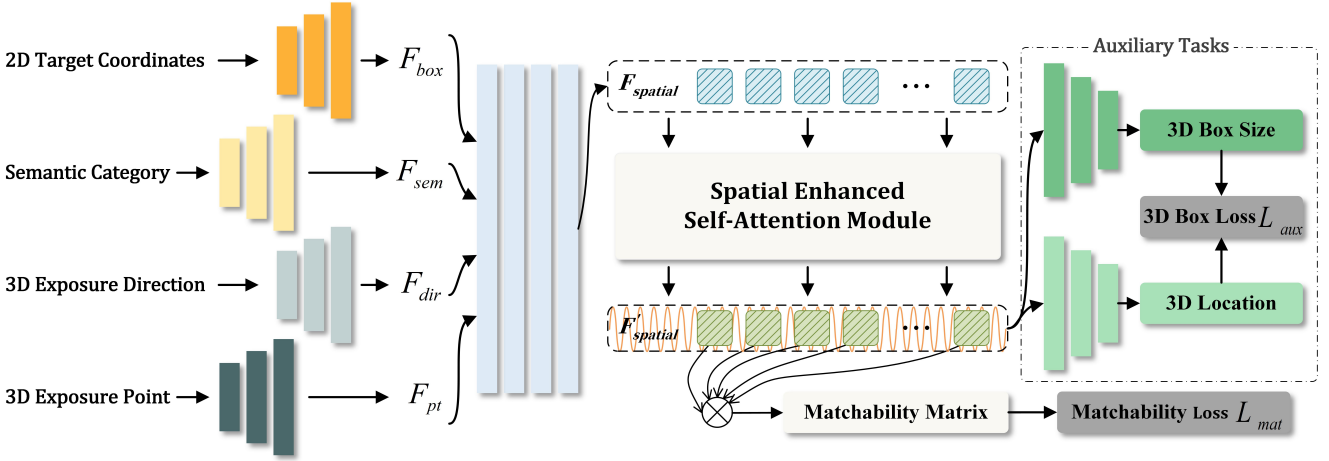}}%
                \caption{Architecture of the spatial attention-based matching module. Each 2D observation is encoded with geometric cues and passed through a Transformer to predict pairwise matchabilities. Auxiliary heads further regularize the representations by regressing 3D box size and location.}
                \label{fig_match}
            \end{figure*}

            \subsection{Infrastructure Identification}
                Infrastructure identification aims to extract unique physical assets from sparse image sequences. This process consists of two sequential stages: first, robustly detecting 2D infrastructure candidates in individual frames; and second, associating these observations across multiple views to assign a unique identity to each physical object.
            \subsubsection{LoRA fine-tuned detection}            
                To enable robust infrastructure detection under limited annotation, we adopt the open-vocabulary detector Grounding DINO \citep{liu2024grounding} and perform lightweight fine-tuning using LoRA (Low-Rank Adaptation) \citep{hu2022lora}.
                This design allows our system to effectively recognize diverse infrastructure objects with minimal manual labeling effort.
                
                Grounding DINO \citep{ren2024grounding} is a transformer-based open-vocabulary detector that supports flexible object localization through cross-modal alignment between image regions and text queries.
                However, directly fine-tuning the entire Grounding DINO model is computationally expensive and prone to overfitting when annotated samples are limited.
    
                To address this, we utilize LoRA, a parameter-efficient fine-tuning technique that introduces trainable low-rank matrices into specific layers of the Transformer (e.g., query and value projections).
                Given a weight matrix $W_0 \in \mathbb{R}^{d \times k}$ in the self-attention module, LoRA decomposes its adaptation into two smaller matrices $A \in \mathbb{R}^{r \times d}$ and $B \in \mathbb{R}^{k \times r}$, where $r \ll d,k$.
                Instead of updating $W_0$ directly, LoRA keeps it frozen and adds a low-rank residual:
                \begin{equation}
                W = W_0 + \Delta W = W_0 + B A.
                \end{equation}
                
                This formulation enables LoRA to achieve fine-grained adaptation by only training $\mathcal{O}(r(d + k))$ additional parameters per projection, while preserving the generalization ability of the original model.
                Furthermore, these low-rank weights can be merged into the backbone during inference without modifying model structure.
                
                Given an input image $\mathcal{I}$ and textual prompts $\mathcal{Q}$, the model outputs a set of detections $\mathcal{B}$:
                \begin{equation}
                \mathcal{B} = \text{Grounding DINO}_{\theta_{\text{LoRA}}}(\mathcal{I}, \mathcal{Q}),
                \end{equation}
                where $\theta_{\text{LoRA}}$ denotes the LoRA-adapted weights, and each detection $b_j \in \mathcal{B}$ is a tuple containing the normalized 2D bounding box center coordinates $(x_j, y_j)$, width $w_j$, height $h_j$, and semantic category label $c_j$.
                

            \subsubsection{Spatial attention-based matching}
                
                In multi-view street imagery, the same infrastructure instance often appears in multiple images due to overlapping fields of view.
                However, as each image lacks depth information, it is difficult to infer accurate 3D positions from individual 2D observations alone.
                This not only limits precise localization but also risks redundant counting of identical objects across views.
                To resolve these issues, we aim to identify and associate observations that correspond to the same physical instance, forming the basis for reliable 3D reasoning.
                
                To establish correspondence across views, we perform matching among detections from each local group of $K$ adjacent panoramic images.
                Let $\mathcal{O} = \{o_i\}_{i=1}^N$ denote the set of $N$ candidate observations within this local window.
                Each observation $o_i$ is represented as a tuple:
                \begin{equation}
                o_i = (p_i, d_i, w_i, h_i, c_i),
                \end{equation}
                where $p_i \in \mathbb{R}^3$ is the 3D exposure position of the camera (derived from GPS or other geolocation data), $w_i$ and $h_i$ are the normalized width and height of the 2D bounding box, and $c_i$ denotes the semantic category label of the observation. The component $d_i \in \mathbb{R}^3$ is the unit vector pointing from the exposure position to the detected object in the world coordinate system, computed as follows:
                \begin{equation}
                d_i = \frac{R_{\text{camera}} \cdot d_{\text{camera}}}{\|R_{\text{camera}} \cdot d_{\text{camera}}\|},
                \end{equation}
                where the direction vector in the camera coordinate system is given by:
                \begin{equation}
                d_{\text{camera}} = 
                \begin{bmatrix}
                \cos(\phi_{\text{box}}) \cos(\theta_{\text{box}}) \\
                \cos(\phi_{\text{box}}) \sin(\theta_{\text{box}}) \\
                \sin(\phi_{\text{box}})
                \end{bmatrix},
                \end{equation}
                with the azimuth angle $\theta_{\text{box}}$ and elevation angle $\phi_{\text{box}}$ defined as:
                \begin{align}
                \theta_{\text{box}} &= \left( \frac{x_j}{\text{image\_width}} \right) \cdot 2\pi - \pi, \\
                \phi_{\text{box}} &= \left( 1 - \frac{y_j}{\text{image\_height}} \right) \cdot \pi - \frac{\pi}{2},
                \end{align}
                where $(x_j, y_j)$ are the normalized bounding box center coordinates of the detection $b_j \in \mathcal{B}$, 
                and $R_{\text{camera}}$ is the rotation matrix from the camera coordinate system to the world coordinate system, parameterized by the Euler angles (heading, pitch, and roll) of the camera, defined as:
                \begin{equation}
                R_{\text{camera}} = R(h_{\text{camera}}, p_{\text{camera}}, r_{\text{camera}}),
                \end{equation}
                where $h_{\text{camera}}$, $p_{\text{camera}}$, and $r_{\text{camera}}$ are the heading, pitch, and roll angles, respectively, obtained from the camera's orientation data, and $R(h_{\text{camera}}, p_{\text{camera}}, r_{\text{camera}})$ represents the composite rotation matrix following the sequence of heading (around Z-axis), pitch (around Y-axis), and roll (around X-axis).
                
                We design a spatial attention-based Transformer $\mathcal{T}_\phi$ \citep{vaswani2017attention} that processes the set $\mathcal{O}$ and outputs a pairwise matchability matrix $S \in \mathbb{R}^{N \times N}$:
                \begin{equation}
                S = \mathcal{T}_\phi(\mathcal{O}),
                \end{equation}
                where each entry $S_{ij} \in [0,1]$ denotes the predicted matching confidence between $o_i$ and $o_j$.
    
                The model attends over the spatial geometric information of observations, learning to associate detections that are spatially consistent across viewpoints.
                As illustrated in Fig.~\ref{fig_match}, each observation is characterized by four types of inputs: the 3D exposure point, the observation direction, the 2D bounding box size, and the semantic category.
                Each of these components is first encoded by a dedicated feature extractor, resulting in the corresponding feature embeddings: $F_{\text{pt}}$, $F_{\text{dir}}$, $F_{\text{box}}$, and $F_{\text{sem}}$, respectively.
                These embeddings are then fused into a unified representation through a feature aggregation module, which produces a token sequence $F_{\text{spatial}}$, capturing the geometric context of all observations.
                
                To enhance mutual awareness among detections and leverage spatial context across views, we introduce a Spatial Enhanced Self-Attention Module. This module takes the fused representation $F_{\text{spatial}}$ as input and enables rich information exchange among tokens, allowing the model to reason over inter-observation relationships and disambiguate spatially proximate but distinct instances. The resulting enhanced token features, denoted as $F'_{\text{spatial}}$, encapsulate contextual dependencies across views and are subsequently fed into a matchability prediction head, where the pairwise matchability between tokens is computed using dot product attention.
                The resulting scores are further passed through a sigmoid activation function to normalize the values into the range $[0,1]$, yielding the final matchability matrix $M$.
                
                Beyond pairwise matching, we enhance the model's spatial awareness by incorporating auxiliary prediction heads that regress the physical size and 3D center of each observed object. These estimations are derived from the updated token features $F'_{\text{spatial}}$, enabling the network to reason about the absolute spatial configuration of observations. By explicitly modeling 3D geometry, the auxiliary tasks help the matching module better distinguish objects that may be close in image space but distant in the real world.

                During training, we jointly optimize the matchability prediction and auxiliary regression objectives. For the matchability matrix $M$, we adopt a binary cross-entropy loss:
                \begin{equation}
                \mathcal{L}_{\text{mat}} = -\frac{1}{N^2} \sum_{i,j} \left[ y_{ij} \log S_{ij} + (1 - y_{ij}) \log (1 - S_{ij}) \right],
                \end{equation}
                where $y_{ij} = 1$ indicates that observations $o_i$ and $o_j$ correspond to the same real-world instance, and $y_{ij} = 0$ otherwise.
                
                For the auxiliary heads, we use L1 loss to supervise the predicted 3D center $\hat{c}_i$ and physical size $\hat{s}_i$ of each observation:
                \begin{equation}
                \mathcal{L}_{\text{aux}} = \frac{1}{N} \sum_{i=1}^{N} \left( \| \hat{c}_i - c_i \|_1 + \| \hat{s}_i - s_i \|_1 \right),
                \end{equation}
                where $c_i$ and $s_i$ are the ground truth 3D center and size, respectively.
                
                The final loss is a weighted sum of the two objectives:
                \begin{equation}
                \mathcal{L} = \mathcal{L}_{\text{mat}} + \lambda \mathcal{L}_{\text{aux}},
                \end{equation}
                where $\lambda$ balances the contribution of the auxiliary task.
                
                By discarding appearance features, our approach is not only more robust to visual ambiguities and domain shifts, but also more lightweight and computationally efficient, with fewer parameters and faster inference.

                To obtain pairwise associations across adjacent frames, we apply a refined version of the Hungarian algorithm to the predicted matchability matrix $M$. For each pair of frames within the $K$-frame local window, we construct a bipartite graph where edge weights are defined by the corresponding entries in $M$, and solve for the optimal one-to-one assignment.
                Since some objects may not appear in all frames, certain observations may lack valid counterparts in adjacent views. To avoid introducing erroneous associations, we discard matches whose confidence falls below a threshold $\tau$, ensuring that only reliable correspondences are retained.
            
            \subsection{Geometry-guided 3D localization}
                Given cross-view observations grouped into preliminary clusters, we perform precise 3D localization through a two-stage geometry-guided process: (1) energy-driven center estimation that minimizes the total point-to-ray distance across all observations in a cluster, and (2) matching refinement that leverages geometric consistency to correct structural errors—specifically resolving over-matching and under-matching—before finalizing the 3D center. This integrated strategy ensures both robustness to local matching noise and high-fidelity localization under sparse imaging conditions.
                \subsubsection{3D center estimation driven by energy optimization}
                    Given a set of matched observations referring to the same object instance, each associated with a known exposure position and a corresponding observation direction, we can geometrically localize the underlying 3D instance center.
                    Specifically, each 2D detection defines a ray in 3D space, originating from its exposure position and extending in the viewing direction toward the object.
                    With multiple such rays from different viewpoints, we estimate the true 3D center of the object as the point that lies closest to all rays, minimizing the total point-to-ray distance.
    
                    Formally, given a cluster $\mathcal{C} = \{(p_i, d_i)\}_{i=1}^{n}$ consisting of $n$ observations, where each observation is represented by a ray with origin $p_i \in \mathbb{R}^3$ and unit direction $d_i \in \mathbb{R}^3$, we define the energy function as the sum of squared distances from the center $c$ to all rays in the cluster:
                    
                    \begin{equation}
                    E(c) = \sum_{i=1}^{n} \left| (c - p_i) - ((c - p_i) \cdot d_i) d_i \right|_2^2
                    \end{equation}
                    
                    Here, the term inside the norm represents the shortest vector from the point $c$ to the ray $(p_i, d_i)$. The optimal 3D center $c^*$ is the point that minimizes this energy function:
                    
                    \begin{equation}
                    c^* = \arg\min_{c \in \mathbb{R}^3} E(c)
                    \label{eq:optimal_center}
                    \end{equation}
                    
                    The energy function reflects how well the point $c$ is positioned relative to the rays. By minimizing this energy, we achieve a geometrically consistent 3D localization.

                \vspace{0.5em}
                \setlength{\algomargin}{0.5em}
                \begin{algorithm}[!t]
                \caption{Geometry-Guided Matching Refinement}
                \label{alg:refinement}
                
                \KwIn{Initial observation clusters $\mathcal{C} = \{C_1, C_2, \dots\}$, where each cluster $C_i = \{o_j\}_{j=1}^{n_i}$ is a set of observations, and each observation $o_j = (p_j, d_j)$ consists of a camera exposure position $p_j \in \mathbb{R}^3$ and a unit observation direction $d_j \in \mathbb{R}^3$\\
                      \hspace{3em}$\tau_{\text{split}}$, $\tau_{\text{merge}}$: distance thresholds for splitting and merging}
                \KwOut{Refined clusters $\mathcal{C}'$}
                
                \vspace{1.0em}
                \textbf{Step 1: Resolve over-matching by splitting inconsistent rays}\\
                Initialize $\mathcal{C}_{\text{temp}} \gets \emptyset$\;
                \For{each cluster $C \in \mathcal{C}$}{
                    \If{$|C| \geq 2$}{
                        Compute 3D center $c^*$ by minimizing total point-to-ray distance (Eq.~\ref{eq:optimal_center})\;
                        Remove all $o = (p, d) \in C$ with $\text{dist}(c^*, \text{ray}(p, d)) > \tau_{\text{split}}$\;
                    }
                    Add updated $C$ to $\mathcal{C}_{\text{temp}}$\;
                }
        
                \vspace{1.0em}
                \textbf{Step 2: Resolve under-matching by recovering missed links}\\
                Let $\mathcal{S} = \{C \in \mathcal{C}_{\text{temp}} \mid |C| = 1\}$, $\mathcal{M} = \{C \in \mathcal{C}_{\text{temp}} \mid |C| \geq 2\}$\;
                \For{each singleton $s = \{(p_s, d_s)\} \in \mathcal{S}$}{
                    Find nearest $m \in \mathcal{M}$ minimizing point-to-ray distance from ray $(p_s, d_s)$ to $c^*_m$\;
                    \If{this distance $< \tau_{\text{merge}}$}{
                        Assign $s$ to $m$\;
                    }
                }
                \For{each pair of singletons $s_i = \{(p_i, d_i)\}, s_j = \{(p_j, d_j)\} \in \mathcal{S}$}{
                    Compute triangulated center $c_{ij}$\;
                    \If{$\text{dist}(c_{ij}, \text{ray}(p_i, d_i)) < \tau_{\text{merge}}$ \textbf{and} $\text{dist}(c_{ij}, \text{ray}(p_j, d_j)) < \tau_{\text{merge}}$}{
                        Estimate physical sizes $\hat{S}_i, \hat{S}_j$ at $c_{ij}$ (Eq.~\ref{eq:scale_check})\;
                        \If{$\max(\frac{\hat{S}_i}{\hat{S}_j}, \frac{\hat{S}_j}{\hat{S}_i}) < \tau_{\text{scale}}$}{
                            Merge $s_i, s_j$ into a new cluster\;
                        }
                    }
                }
                
                \vspace{1.0em}
                \textbf{Step 3: Finalize refined clusters}\\
                Recompute 3D centers for all updated clusters using Eq.~\ref{eq:optimal_center}\;
                \Return $\mathcal{C}'$\;
                
                \end{algorithm}
                    
                    \textbf{Initialization}: The triangulation process begins by averaging pairwise ray intersections in the XY plane to estimate the horizontal position of the object. The vertical coordinate is computed by projecting each 2D intersection back onto its corresponding 3D ray and averaging the resulting Z values. This initialization ensures that the estimated center is close to the actual ray intersections in 3D space.
                    
                    \textbf{Refinement}: To improve the accuracy of this initial estimate, we apply the Broyden–Fletcher–Goldfarb–Shanno (BFGS) algorithm \citep{head1985broyden}, a quasi-Newton method that approximates the inverse Hessian matrix using first-order gradients. The optimization further minimizes the energy function, refining the 3D localization result by adjusting the center point to fit all the rays more consistently.
                    
                \subsubsection{Geometry-guided matching refinement}

                    Although the global geometric optimization is robust to minor matching noise—since the estimated 3D center $c^*$ (Eq.~\eqref{eq:optimal_center}) is dominated by geometrically consistent rays—it remains highly sensitive to structural matching errors. Specifically, \textbf{over-matching} (i.e., merging observations from distinct physical instances) violates the single-instance geometric assumption, leading to a biased center and inflated asset counts; conversely, \textbf{under-matching} (i.e., fragmented observations of the same instance) results in insufficient or poorly conditioned ray sets, degrading triangulation accuracy or rendering it infeasible.

\begin{figure}[!htbp]
                    \centering
                    \includegraphics[width=0.45\textwidth]{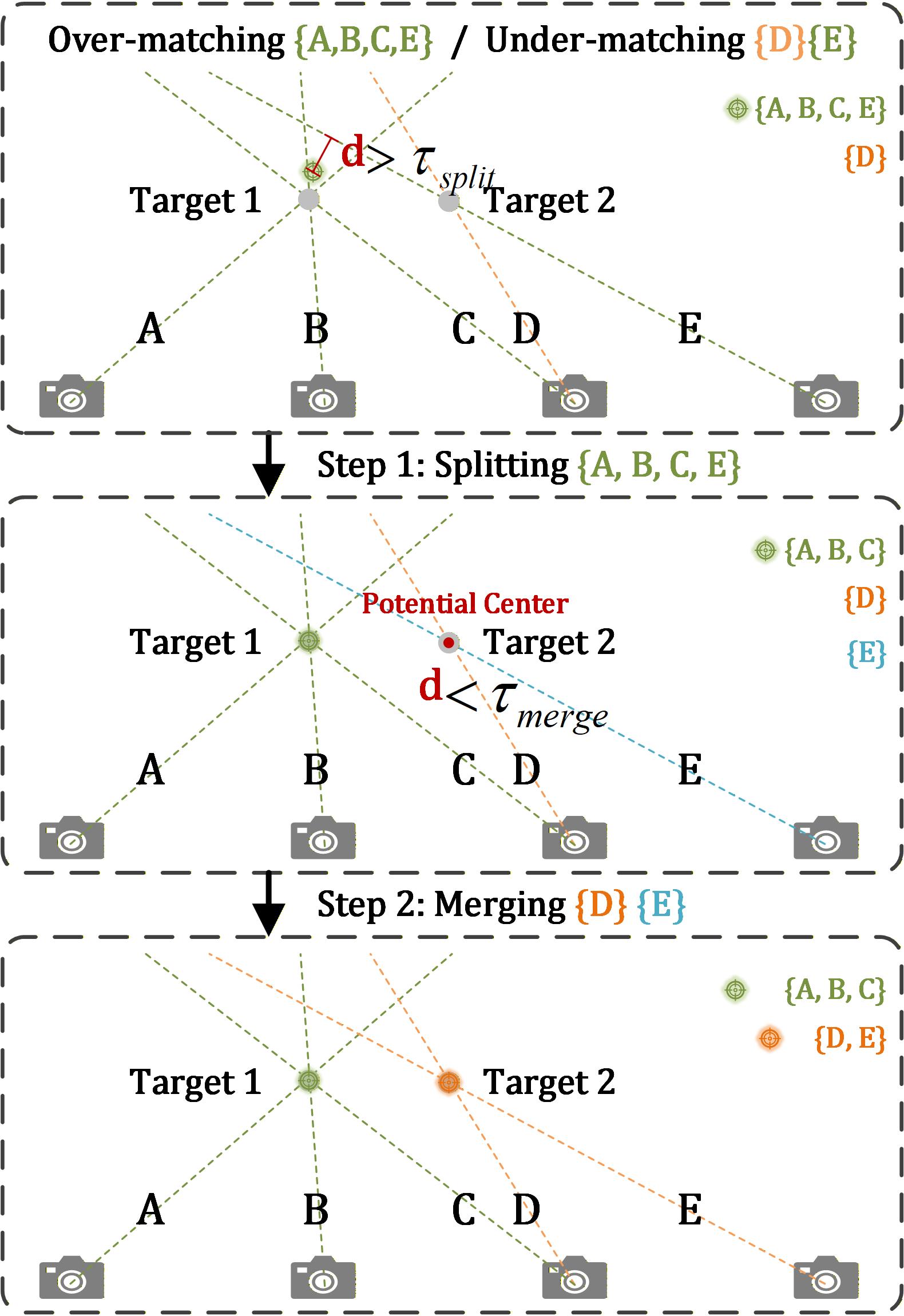}
                    \caption{Illustration of the geometry-guided matching refinement process. \textbf{Note that for visualization clarity, the ray-to-center relationships are simplified in 2D, whereas our actual algorithm operates in 3D space.}
                    (Top) \textbf{Initial State:} Due to erroneous matching, ray $E$ (belonging to Target 2) is grouped with Target 1, causing a large geometric residual ($d > \tau_{split}$). 
                    (Middle) \textbf{Step 1: Splitting.} Ray $E$ is identified as a geometric outlier and pruned from the cluster, correcting the center of Target 1. 
                    (Bottom) \textbf{Step 2: Merging.} The isolated ray $E$ is re-evaluated and merged with ray $D$ based on geometric consistency ($d < \tau_{merge}$), successfully recovering Target 2.}
                    \label{fig:match_refine}
                \end{figure}

                    To address these issues, we propose a \textbf{geometry-guided matching refinement} strategy (Algorithm~\ref{alg:refinement}) that leverages 3D triangulation both to prune inconsistent associations and to recover plausible missed links. 
                    \textbf{For intuitive understanding, Fig.~\ref{fig:match_refine} visualizes the core mechanism using a simplified 2D projection, although the actual algorithm operates in 3D space.} 

                    Starting from the initial matched graph (Fig.~\ref{fig:match_refine}, Top), we first estimate a 3D center $c^*$ for each connected component using Eq.~\eqref{eq:optimal_center}. Observations whose point-to-ray distances exceed a category-adaptive threshold $\tau_{\text{split}}$ are pruned. As shown in the transition to the \textbf{Middle panel} of Fig.~\ref{fig:match_refine}, this effectively resolves over-matching by removing geometric outliers (e.g., ray $E$).
                    
                    Subsequently, we examine unassigned or singleton observations for potential merges: a singleton is absorbed into an existing multi-view cluster if its ray-to-center distance is below $\tau_{\text{merge}}$; pairwise singletons are merged only if both rays lie within $\tau_{\text{merge}}$ of the center and satisfy a physical size consistency check. 
                    Since ray intersection alone can yield false positives (e.g., coincidental crossing of rays from unrelated objects), we verify that the observations imply a consistent physical scale. Specifically, for an observation with exposure position $p \in \mathbb{R}^3$, ray direction $d \in \mathbb{R}^3$, and 2D bounding box size $s_{\text{2d}}$ (height or width), we estimate its physical size $\hat{S}$ at the triangulated center $c_{\text{tri}}$ as:
                    \begin{equation}
                    \hat{S} \approx s_{\text{2d}} \cdot \left| (c_{\text{tri}} - p) \cdot d \right|.
                    \label{eq:scale_check}
                    \end{equation}
                    The term $|(c_{\text{tri}} - p) \cdot d|$ represents the projection depth along the viewing ray. The merge is executed only if the estimated sizes from two views $i$ and $j$ are consistent, i.e., $\max(\frac{\hat{S}_i}{\hat{S}_j}, \frac{\hat{S}_j}{\hat{S}_i}) < \tau_{\text{scale}}$. This bidirectional refinement enhances both precision (by suppressing geometrically incompatible merges) and recall.

                    Finally, we re-estimate $c^*$ for all refined clusters using Eq.~\eqref{eq:optimal_center}. The resulting centers are computed from geometrically coherent ray sets, ensuring centimeter- to decimeter-level localization accuracy and faithful instance counting, even under sparse and noisy viewing conditions.
                
            \subsection{State discriminative VLM agent}

                In infrastructure perception and management, relying solely on geometric or semantic recognition is insufficient for fine-grained operation and maintenance. Attribute characterization provides accurate static descriptions of objects, while state discrimination reflects their functional and operational conditions. Together, they not only enrich semantic representation but also offer essential evidence for prioritizing maintenance, detecting anomalies, and analyzing spatiotemporal changes.
                
\begin{figure}[!h]
                \centering{
                \includegraphics[width=0.45\textwidth]{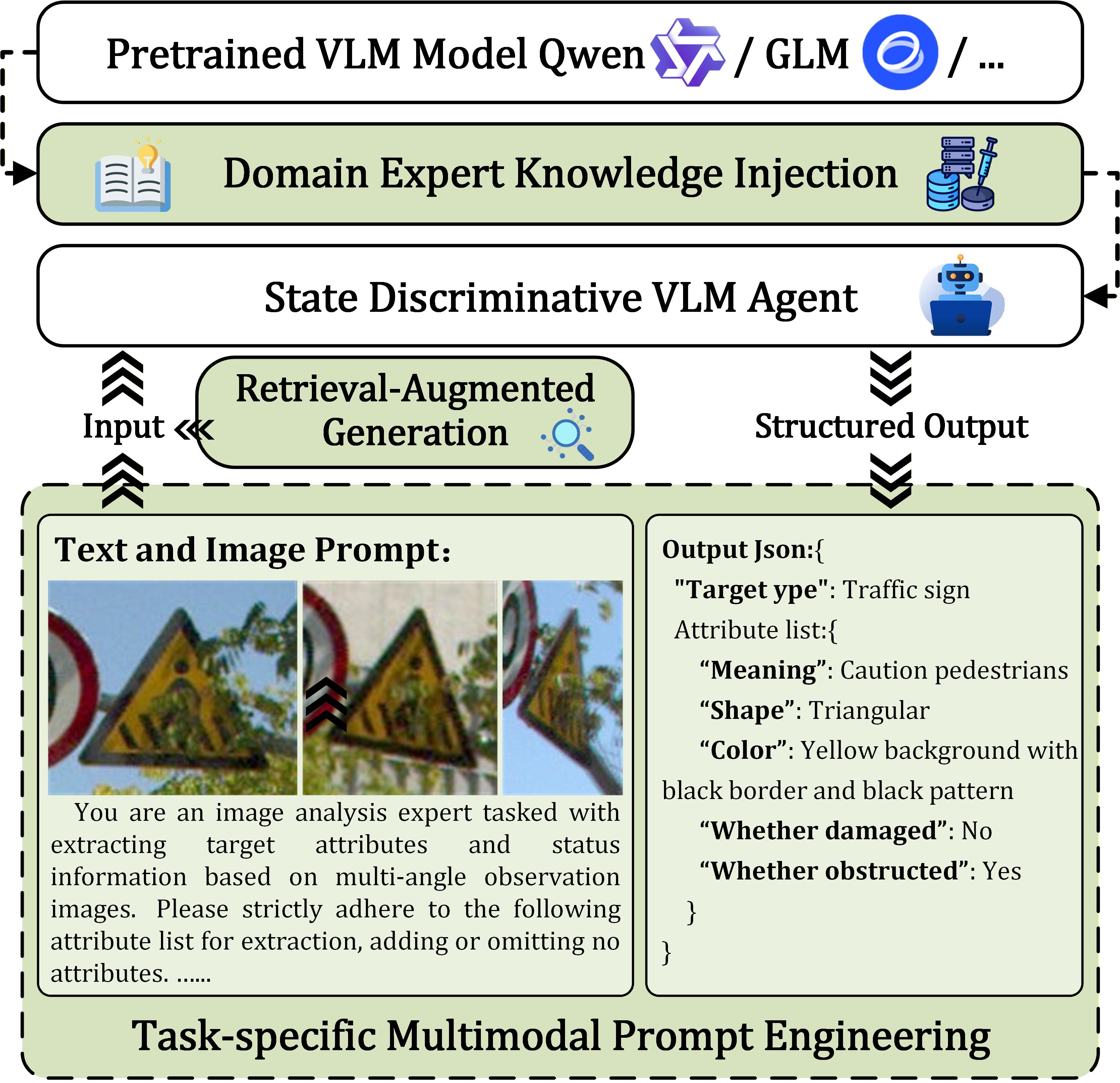}}%
                \caption{Workflow of the State-discriminative VLM Agent. The agent integrates Expert Knowledge Injection and Retrieval-Augmented Generation (RAG) to enhance a pre-trained VLM. Through Task-specific Multimodal Prompting with multi-view images, it accurately infers fine-grained attributes and operational states (e.g., "Obstructed"), delivering results in a structured JSON format.}
                \label {fig_vlm_agent}
                \end{figure}

                To achieve this fine-grained perception (e.g., reliably distinguishing structural damage from surface dirt), we design a State-discriminative VLM Agent.
                This agent utilizes a pre-trained large vision-language model (e.g., Qwen-VL \citep{bai2023qwen, wang2024qwen2}, GLM-4v \citep{glm2024chatglm}, or LLaVA \citep{liu2023visual}) as its cognitive backbone, tasked with diagnosing the detailed attributes and operational health status of the identified infrastructure targets from the associated multi-view images.
                As shown in Fig.~\ref{fig_vlm_agent}, rather than fine-tuning the large model—which is computationally expensive and risks catastrophic forgetting—we adopt a \textbf{training-free inference strategy}. This framework guides the model's reasoning through three synergistic modules: Task-specific Multi-modal Prompting, Expert Knowledge Injection, and Retrieval-Augmented Generation (RAG) \citep{lewis2020retrieval}.
                
                \textbf{Task-specific multi-modal prompting.}
                Generic prompts often yield irrelevant captions. To enforce domain focus, we construct a structured prompt containing: (i) \textbf{Cropped ROI images} $\{I_j\}$ to isolate the target; (ii) \textbf{Global context view} to provide environmental cues (e.g., lighting); and (iii) \textbf{Explicit task instructions}. For instance, instead of asking for a general description, we instruct the model to "Check if the pole is tilted" or "Count the lamp heads," effectively formulating the task as a constrained reasoning problem.
                
                \textbf{Expert Knowledge Injection.}
                However, general reasoning is insufficient without specific engineering criteria. We bridge this gap by injecting a structured rulebook derived from national standards and expert maintenance manuals. We define specific \textbf{symptom-to-state mapping rules} to guide inference. For instance, in the case of \textit{Street Lights}, a visual symptom of "dark lamp with intact housing" is explicitly mapped to the state of \textbf{Lamp Failure}. Similarly, for \textit{Traffic Signs}, we encode a degradation hierarchy where "legible but faded color" indicates \textbf{Minor Degradation}, whereas a "missing panel" corresponds to \textbf{Critical Damage}. These rules act as a logic filter, ensuring the model's output aligns with professional maintenance categories.
                
                \textbf{Retrieval augmentation (RAG).}
                While expert rules provide logical definitions, \textbf{textual descriptions alone are often insufficient to define subtle visual states} (e.g., differentiating "faded paint" from "dirt coverage"). We address this limitation by providing \textbf{visual references}. We maintain a database $\mathcal{D}$ of annotated exemplars. For a query $q$, we retrieve the top-$K$ visually similar examples:
                \begin{equation}
                \mathcal{R}_K(q) = \mathrm{TopK}_{(x_i, y_i) \in \mathcal{D}} \frac{f(q) \cdot f(x_i)}{\|f(q)\| \, \|f(x_i)\|}.
                \end{equation}
                These retrieved examples and their labels are inserted into the prompt context. This enables the VLM to perform analogical reasoning, comparing the query against confirmed reference cases to resolve visual ambiguities that text rules cannot capture.
                
                During inference, the relevant subset of this knowledge base and retrieved exemplars are dynamically embedded as a contextual block within the multi-modal prompt. This enables the VLM to perform regulation-aware and symptom-guided reasoning without fine-tuning, significantly improving the consistency, interpretability, and real-world applicability of its state predictions.

            \begin{table}[!h]
                \centering
                \renewcommand{\arraystretch}{1.0}
                \caption{Statistics of infrastructure categories in the \textbf{Evaluation Subsets}. The table reports the number of unique physical objects (Obj.) and the total number of 2D observations (Obs.) used for cross-view matching and geo-localization evaluation.}
                \label{table_insnumber}
                \fontsize{8}{9}\selectfont
                \begin{tabular}{p{2.7cm} >{\centering\arraybackslash}p{1.0cm} >{\centering\arraybackslash}p{1.0cm} >{\centering\arraybackslash}p{1.0cm} >{\centering\arraybackslash}p{1.0cm}}
                    \toprule
                    \multirow{2}{*}{Class} & \multicolumn{2}{c}{Wuhan (Eval. Set)} & \multicolumn{2}{c}{Shanghai (Eval. Set)} \\
                    \cmidrule(lr){2-3} \cmidrule(lr){4-5}
                                           & Obj. & Obs. & Obj. & Obs. \\
                    \midrule
                    Traffic Sign        & 127 & 1334 & 166 & 2134 \\ 
                    Street Light        & 97  & 1236 & 148 & 2593 \\ 
                    Signal Light        & 109 & 1037 & 112 & 1919 \\ 
                    Surveillance Camera & 36  & 293  & 26  & 135  \\ 
                    Barrier Bollard     & 116 & 694 & --  & --   \\ 
                    Safety Bollard      & --  & --  & 198 & 1863 \\ 
                    Fire Hydrant        & 8   & 45   & 19  & 189  \\ 
                    Trash Bin           & 16  & 93   & 40  & 248  \\ 
                    Manhole             & 159 & 454  & 158 & 852  \\ 
                    Traffic Cone        & 182 & 892  & 35  & 180  \\ 
                    Spherical Bollard   & 38  & 190  & 22  & 172  \\ 
                    \bottomrule
                \end{tabular}
            \end{table}
        
        \section{Experiments and results}

            \subsection{Dataset description and experiment settings}
                \textbf{Dataset Construction.} We conduct experiments on two urban road datasets collected in Wuhan and Shanghai using a Mobile Mapping System (MMS). The Wuhan dataset covers a total trajectory of 20.1\,km, while the Shanghai dataset spans 21.8\,km. 
                Although the acquisition system captures synchronized LiDAR point clouds, our framework is designed to be vision-only. This design choice avoids the high costs of equipment, storage, and computation associated with LiDAR, which are often prohibitive for routine infrastructure inventory. \textbf{In this study, the point clouds are utilized strictly to generate high-precision ground truth for verifying our method's accuracy and are not involved in the inference pipeline.}
                
                Since the raw data lacks object-level annotations, we constructed two distinct subsets for each city to serve different stages of our framework:
                
                (1) \textbf{Detection Fine-tuning Subset:} To adapt the open-set detector to roadside infrastructure, we annotated a large-scale subset consisting of 1,746 panoramic images for Wuhan and 1,308 for Shanghai. These images are labeled solely with 2D bounding boxes and category names.
                
                (2) \textbf{Evaluation Subset:} For training the matching network and evaluating the full pipeline, we densely annotated a continuous sequence of 229 images for Wuhan and 290 images for Shanghai. This subset includes 2D bounding boxes and unique observation IDs to link the same physical object across frames. Crucially, we manually annotated 3D bounding boxes on the corresponding point clouds to calculate the geometric center of each object. These centers serve as the ground truth for evaluating the geo-localization accuracy.
                Table~\ref{table_insnumber} summarizes the statistics of infrastructure categories in the \textit{Evaluation Subsets} for both cities.
                
                \textbf{Experimental Protocol.} 
                For object detection, we utilize the corresponding \textit{Detection Fine-tuning Subset} of each city to fine-tune the Grounding DINO model via LoRA.
                For the matching and localization stages, we adopt a \textbf{cross-city evaluation protocol} to rigorously assess the generalization capability of our spatial-attention matching network. Specifically, the matching network is trained on the fully annotated Wuhan evaluation subset (229 images) and tested on the Shanghai evaluation subset (290 images). Conversely, we also report results by training on Shanghai and testing on Wuhan. This setting ensures that the model learns geometric invariance rather than overfitting to specific city layouts or visual appearances.

                \begin{figure*}[!h]
                    \centering
                    \includegraphics[width=0.95\linewidth]{Figs/image_detect.jpg}
                    \caption{Visualization of object detection results on panoramic street-view images using the LoRA fine-tuned Grounding DINO model.}
                    \label{fig_image_detect}
                \end{figure*}
                
            \subsection{Quantitative evaluation}
                
                We quantitatively evaluate the proposed pipeline across three stages: image-level object detection, cross-view target matching, and final 3D identification and geo-localization. For each stage, task-specific metrics are adopted to measure corresponding performance comprehensively. 
                  
                \subsubsection{Object detection}
                    In the object detection phase, we employed the Grounding DINO model fine-tuned via Low-Rank Adaptation (LoRA) to adapt to the specific domain of roadside infrastructure. We evaluated the performance of this fine-tuned detector on both the Wuhan and Shanghai sparse street-view datasets. Table~\ref{table:image_detect} presents the quantitative results. We report the Average Precision at an Intersection-over-Union (IoU) threshold of 0.5 (denoted as AP@0.5) \citep{lin2014microsoft}. Additionally, to provide a detailed view of detection completeness and correctness, we report Precision (Pre$_{\text{(det)}}$) and Recall (Rec$_{\text{(det)}}$), both calculated at the same IoU threshold of 0.5.

                    \begin{table}[!h]
                        \centering
                        \renewcommand{\arraystretch}{1.0}
                        \caption{Detection performance (AP@0.5, Precision, and Recall) after LoRA fine-tuning on the Wuhan and Shanghai datasets.}
                        \label{table:image_detect}
                        \fontsize{8}{9}\selectfont
                        \resizebox{0.50\textwidth}{!}{%
                        \begin{tabular}{lcccccc}
                            \toprule
                            \multirow{2}{*}{Class} & \multicolumn{3}{c}{Wuhan} & \multicolumn{3}{c}{Shanghai} \\
                            \cmidrule(lr){2-4} \cmidrule(lr){5-7}
                            & AP@0.5 & Pre$_{\text{(det)}}$ & Rec$_{\text{(det)}}$ & AP@0.5 & Pre$_{\text{(det)}}$ & Rec$_{\text{(det)}}$ \\
                            \midrule
                            Traffic Sign        & 68.7 & 78.9 & 74.4 & 84.2 & 81.7 & 90.1 \\
                            Street Light        & 81.4 & 63.0 & 89.6 & 79.9 & 84.8 & 88.3 \\
                            Signal Light        & 68.7 & 75.1 & 80.9 & 87.7 & 89.7 & 93.3 \\
                            Surveillance Camera & 65.3 & 67.7 & 74.6 & 54.8 & 58.4 & 68.2 \\
                            Barrier Bollard     & 70.8 & 72.0 & 81.1 & -- & -- & -- \\
                            Safety Bollard      & -- & -- & -- & 89.7 & 87.2 & 92.5 \\
                            Fire Hydrant        & 66.1 & 77.3 & 75.6 & 91.2 & 84.9 & 95.5 \\
                            Trash Bin           & 85.9 & 86.8 & 88.1 & 70.4 & 79.8 & 74.9 \\
                            Manhole             & 62.8 & 67.0 & 71.6 & 54.7 & 59.9 & 69.1 \\
                            Traffic Cone        & 79.6 & 79.3 & 83.4 & 83.1 & 80.4 & 87.3 \\
                            Spherical Bollard   & 81.6 & 77.9 & 86.7 & 79.7 & 94.8 & 83.2 \\
                            \midrule
                            Average             & 73.1 & 74.5 & 80.6 & 77.5 & 80.2 & 84.2 \\
                            \bottomrule
                        \end{tabular}%
                        }
                    \end{table}
                    
                    As shown in Table~\ref{table:image_detect}, the model demonstrates robust detection capabilities despite being trained on limited labeled data. The average AP@0.5 reached 73.1\% on the Wuhan dataset and 77.5\% on the Shanghai dataset, proving the effectiveness of the LoRA strategy in bridging the domain gap between general pre-training data (e.g., COCO) and specific urban scenes.
                    
                    From a category-specific perspective, the detector performs exceptionally well on distinct, upright objects. For instance, `Street Light', `Signal Light', and `Fire Hydrant' (in Shanghai) achieved high scores, with Recall rates often exceeding 88\%. However, performance drops slightly for small or low-texture objects such as `Surveillance Camera' (AP 54.8\%-65.3\%) and `Manhole' (AP 54.7\%-62.8\%). This is primarily due to the small pixel area of distant cameras \citep{cheng2023towards} and the significant perspective distortion of manholes on the ground surface.
                
                \subsubsection{Target matching}
                    Evaluating the quality of cross-view associations is crucial for understanding both the effectiveness of the proposed matching model and the reliability of the final results. 
                    \textbf{Note that to decouple the matching performance from upstream detection errors, we utilize ground-truth 2D bounding boxes as input for both training and testing in this evaluation.} This ensures that the reported metrics purely reflect the association capability of the matching network.
                    To this end, we assess target matching at two different levels. 

                    \begin{table}[!h]
                    \centering
                    \renewcommand{\arraystretch}{1.0}
                    \caption{Local pairwise matching performance (\%) across different categories for the Wuhan and Shanghai dataset. Precision, Recall, and F1-score are computed from the binary matching matrices of three adjacent frames.}
                    \label{table:local_matching}
                    \fontsize{8}{9}\selectfont
                    \resizebox{0.50\textwidth}{!}{
                    \begin{tabular}{lcccccc}
                        \toprule
                        \multirow{2}{*}{Class} & \multicolumn{3}{c}{Wuhan} & \multicolumn{3}{c}{Shanghai} \\
                        \cmidrule(lr){2-4} \cmidrule(lr){5-7}
                        & Pre$_{\text{(mat)}}$ & Rec$_{\text{(mat)}}$ & F1$_{\text{(mat)}}$ & Pre$_{\text{(mat)}}$ & Rec$_{\text{(mat)}}$ & F1$_{\text{(mat)}}$ \\
                        \midrule
                        Traffic Sign        & 0.963 & 0.958 & 0.961 & 0.935 & 0.945 & 0.940 \\
                        Street Light        & 0.993 & 0.993 & 0.993 & 0.971 & 0.968 & 0.969 \\
                        Signal Light        & 0.957 & 0.951 & 0.954 & 0.942 & 0.914 & 0.928 \\
                        Surveillance Camera & 0.962 & 0.962 & 0.962 & 0.907 & 0.907 & 0.907 \\
                        Barrier Bollard     & 0.843 & 0.859 & 0.851 & --    & --    & --    \\
                        Safety Bollard      & --    & --    & --    & 0.733 & 0.705 & 0.719 \\
                        Fire Hydrant        & 1.000 & 1.000 & 1.000 & 1.000 & 1.000 & 1.000 \\
                        Trash Bin           & 1.000 & 1.000 & 1.000 & 1.000 & 1.000 & 1.000 \\
                        Manhole             & 0.908 & 0.928 & 0.918 & 0.919 & 0.928 & 0.924 \\
                        Traffic Cone        & 0.786 & 0.794 & 0.790 & 0.856 & 0.865 & 0.861 \\
                        Spherical Bollard   & 0.893 & 0.900 & 0.896 & 0.899 & 0.875 & 0.887 \\
                        \midrule
                        Average             & 0.931 & 0.935 & 0.933 & 0.916 & 0.911 & 0.913 \\
                        \bottomrule
                    \end{tabular}
                    }
                \end{table}
    
                    At the \textbf{local pairwise level}, given the ground-truth matching matrix $Y \in \{0,1\}^{N \times N}$ and the predicted binary matrix $\hat{Y} \in \{0,1\}^{N \times N}$, where $N$ denotes the number of ground-truth observations from three consecutive frames jointly input to the model, each entry $Y_{ij}=1$ indicates that observations $o_i$ and $o_j$ belong to the same physical object. Based on these matrices, we compute:
                    \begin{equation}
                    \begin{aligned}
                    \text{Pre}_{\text{(mat)}} &= \frac{TP_{\text{(mat)}}}{TP_{\text{(mat)}}+FP_{\text{(mat)}}}, \\
                    \text{Rec}_{\text{(mat)}} &= \frac{TP_{\text{(mat)}}}{TP_{\text{(mat)}}+FN_{\text{(mat)}}}, \\
                    \text{F1}_{\text{(mat)}} &= \frac{2 \cdot \text{Pre}_{\text{(mat)}} \cdot \text{Rec}_{\text{(mat)}}}{\text{Pre}_{\text{(mat)}} + \text{Rec}_{\text{(mat)}}},
                    \end{aligned}
                    \label{eq:match_metrics}
                    \end{equation}
                    where $TP_{\text{(mat)}} = \sum_{i,j} \mathbb{I}(\hat{Y}_{ij}=1 \wedge Y_{ij}=1)$, $FP_{\text{(mat)}} = \sum_{i,j} \mathbb{I}(\hat{Y}_{ij}=1 \wedge Y_{ij}=0)$, and $FN_{\text{(mat)}} = \sum_{i,j} \mathbb{I}(\hat{Y}_{ij}=0 \wedge Y_{ij}=1)$, $\mathbb{I}(\cdot)$ is the indicator function which equals 1 if the condition is true and 0 otherwise.
                    These metrics directly reflect the accuracy of local association decisions by the proposed matching model.
                    As shown in Table~\ref{table:local_matching}, our method achieves robust pairwise association, with average F1-scores of 93.3\% and 91.3\% on the Wuhan and Shanghai datasets, respectively. High precision and recall scores are consistently observed across diverse categories—ranging from tall \textit{Street Lights} (F1 $>96$\%) to ground-level \textit{Manholes} (F1 $>91$\%). This indicates that the proposed spatial-attention mechanism effectively captures the geometric consistency required to link observations across wide-baseline frames, regardless of object elevation or scale.

                    At the \textbf{global clustering level}, let $y = (y_1, \dots, y_{N_{\text{all}}})$ denote the ground-truth object labels and $c = (c_1, \dots, c_{N_{\text{all}}})$ the predicted cluster assignments for all $N_{\text{all}}$ observations. We report \textit{Homogeneity}, \textit{Completeness}, and their harmonic mean \textit{V-Measure} \citep{rosenberg2007v}, defined as:
                    \begin{equation}
                    \begin{aligned}
                    \text{Homogeneity} &= 1 - \frac{H(y\,|\,c)}{H(y)}, \\
                    \text{Completeness} &= 1 - \frac{H(c\,|\,y)}{H(c)}, \\
                    \text{V\text{-}Measure} &= \frac{2 \cdot \text{Homogeneity} \cdot \text{Completeness}}{\text{Homogeneity} + \text{Completeness}},
                    \end{aligned}
                    \label{eq:global_metrics}
                    \end{equation}
                    where $H(\cdot)$ and $H(\cdot|\cdot)$ denote entropy and conditional entropy, which are defined as:
                    \begin{equation}
                    \begin{aligned}
                    H(y) &= - \sum_{m} \frac{n_m}{N_{\text{all}}}\,\log \frac{n_m}{N_{\text{all}}}, \\
                    H(c) &= - \sum_{k} \frac{n_k}{N_{\text{all}}}\,\log \frac{n_k}{N_{\text{all}}}, \\
                    H(y\,|\,c) &= - \sum_{k} \frac{n_k}{N_{\text{all}}} \sum_{m} \frac{n_{km}}{n_k}\,\log \frac{n_{km}}{n_k}, \\
                    H(c\,|\,y) &= - \sum_{m} \frac{n_m}{N_{\text{all}}} \sum_{k} \frac{n_{km}}{n_m}\,\log \frac{n_{km}}{n_m}.
                    \end{aligned}
                    \label{eq:entropy_defs}
                    \end{equation}
                    where $n_{km} = |C_k \cap G_m|$ is the number of observations assigned to predicted cluster $C_k$ with ground-truth label $G_m$, $n_k=\sum_m n_{km}$, $n_m=\sum_k n_{km}$, and $N_{\text{all}}=\sum_{k,m} n_{km}$. 
    
                    Homogeneity measures the extent to which each predicted cluster contains only observations of a single ground-truth object, while Completeness measures whether all observations of a given ground-truth object are assigned to the same cluster. Together with V-Measure, these metrics evaluate the agreement between predicted clusters and ground-truth identities. 

                    \begin{table}[!h]
                    \centering
                    \renewcommand{\arraystretch}{1.0}
                    \caption{Homogeneity (Homo.), Completeness (Comp.), and V-Measure (V-Meas.) (\%) across different categories for the Wuhan dataset and Shanghai dataset.}
                    \label{table:global_matching}
                    \fontsize{8}{9}\selectfont
                    \resizebox{0.50\textwidth}{!}{%
                    \begin{tabular}{lcccccc}
                        \toprule
                        \multirow{2}{*}{Class} & \multicolumn{3}{c}{Wuhan} & \multicolumn{3}{c}{Shanghai} \\
                        \cmidrule(lr){2-4} \cmidrule(lr){5-7}
                        & Homo. & Comp. & V-Meas. & Homo. & Comp. & V-Meas. \\
                        \midrule
                        Traffic Sign        & 0.997 & 0.977 & 0.987 & 0.994 & 0.978 & 0.986 \\
                        Street Light        & 0.996 & 0.994 & 0.995 & 0.990 & 0.981 & 0.986 \\
                        Signal Light        & 0.979 & 0.975 & 0.977 & 0.985 & 0.976 & 0.981 \\
                        Surveillance Camera & 0.995 & 0.992 & 0.993 & 0.978 & 0.978 & 0.978 \\
                        Barrier Bollard     & 0.997 & 0.959 & 0.978 & --    & --    & --    \\
                        Safety Bollard      & --    & --    & --    & 0.984 & 0.954 & 0.969 \\
                        Fire Hydrant        & 1.000 & 1.000 & 1.000 & 1.000 & 0.992 & 0.996 \\
                        Trash Bin           & 1.000 & 0.995 & 0.997 & 1.000 & 0.997 & 0.999 \\
                        Manhole             & 0.999 & 0.960 & 0.979 & 0.994 & 0.981 & 0.987 \\
                        Traffic Cone        & 0.971 & 0.919 & 0.944 & 0.996 & 0.939 & 0.967 \\
                        Spherical Bollard   & 0.994 & 0.982 & 0.988 & 0.971 & 0.971 & 0.971 \\
                        \midrule
                        Average             & 0.993 & 0.975 & 0.984 & 0.989 & 0.975 & 0.982 \\
                        \bottomrule
                    \end{tabular}
                    }
                \end{table}

            \begin{figure*}[!h]
            \centering{
            \includegraphics[width=0.95\textwidth]{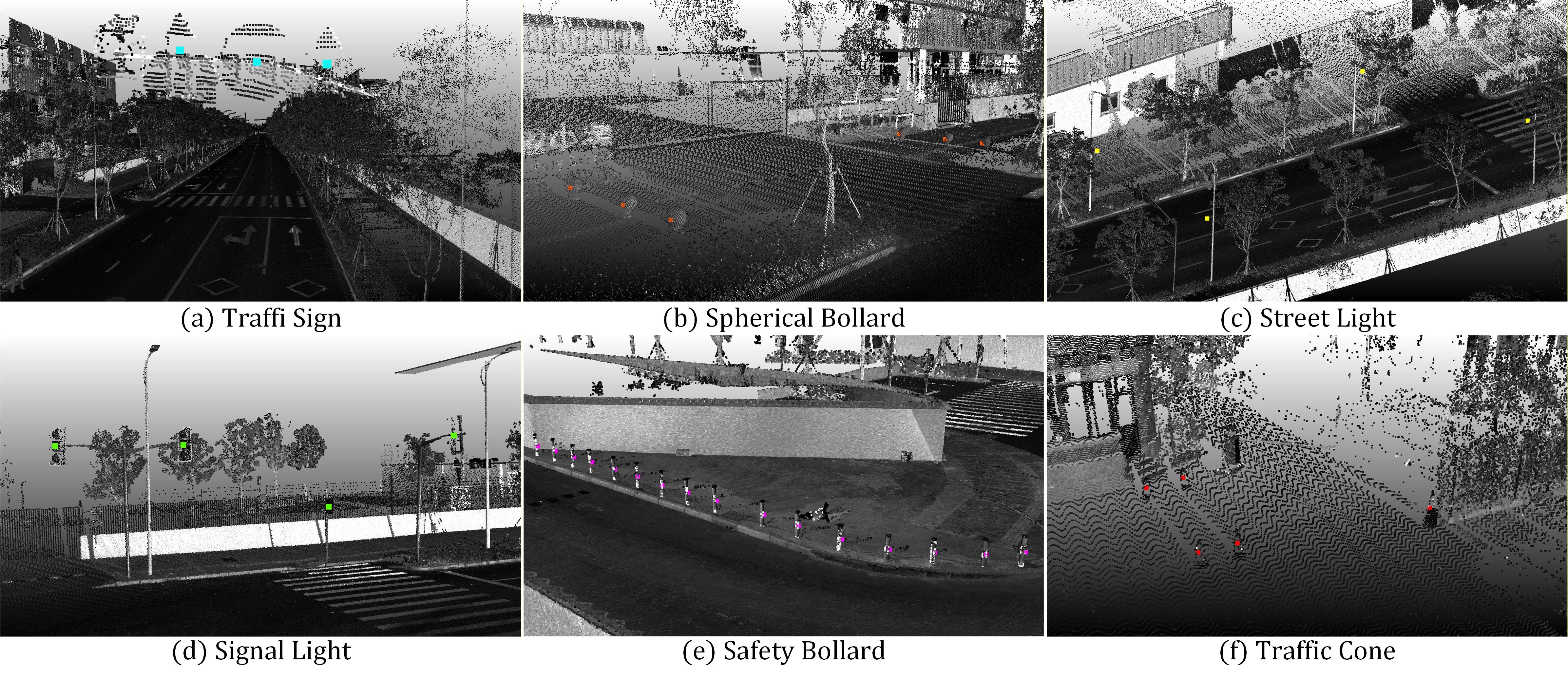}}%
            \caption{Visualizations of extracted 3D centers for various categories, overlaid on the corresponding point clouds. Each colored marker represents the estimated center, demonstrating accurate localization for categories such as (a) Traffic Signs, (b) Spherical Bollards, (c) Street Lights, (d) Signal Lights, (e) Safety Bollards, and (f) Traffic Cones.}
            \label {fig_detail_categories}
            \end{figure*}
                    
                    Table~\ref{table:global_matching} validates the high quality of our global clustering, with average V-Measures surpassing 0.98 for both cities. Categories like \textit{Fire Hydrants} achieve perfect scores primarily due to their high spatial sparsity; since multiple instances rarely co-occur within the same local window, the ambiguity in association is minimal. 
                    Crucially, for visually repetitive and dense categories like \textit{Spherical Bollards}, our method maintains robust performance (V-Measure $>0.98$ in Wuhan and $>0.97$ in Shanghai). This success is attributed to the integration of 3D geometric cues, which allow the model to distinguish visually identical but spatially distinct instances—a common failure case for appearance-based trackers.
                    
                    Furthermore, under the challenging cross-city evaluation setting—where models are trained on one city and tested on the other—our method achieves consistent high performance (average V-Measure $>0.98$ for both datasets). This robustness against domain shifts in urban layouts suggests that the proposed spatial-attention mechanism successfully learns intrinsic geometric invariance, rather than overfitting to the specific visual characteristics of the training city, holding great potential for scalable deployment in unseen cities.
                
                \subsubsection{3D identification and geo-localization}
                    In contrast to the standalone matching evaluation which uses ground-truth boxes, the results reported here are obtained from the full inference pipeline, utilizing predicted bounding boxes from our detection module. This reflects the system's performance in real-world deployment.
                    
                    To evaluate the system's performance in both \textbf{target identification} and \textbf{3D localization}, we consider a prediction to be correct if the distance between its predicted center $\hat{p}_i \in \mathbb{R}^3$ and the ground-truth center $p_i \in \mathbb{R}^3$ is less than $1\,\text{m}$. Based on this criterion, we compute:
                    \begin{equation}
                    \begin{aligned}
                    \text{Pre}_{\text{(idf)}} &= \frac{TP_{\text{(idf)}}}{TP_{\text{(idf)}}+FP_{\text{(idf)}}}, \\
                    \text{Rec}_{\text{(idf)}} &= \frac{TP_{\text{(idf)}}}{TP_{\text{(idf)}}+FN_{\text{(idf)}}}, \\
                    \text{F1}_{\text{(idf)}} &= \frac{2 \cdot \text{Pre}_{\text{(idf)}} \cdot \text{Rec}_{\text{(idf)}}}{\text{Pre}_{\text{(idf)}} + \text{Rec}_{\text{(idf)}}},
                    \end{aligned}
                    \label{eq:loc_metrics}
                    \end{equation}
                    where $TP_{\text{(idf)}}$ denotes \textbf{successfully identified targets} (predictions matched to a ground truth within $1\,\text{m}$), $FP_{\text{(idf)}}$ represents \textbf{false positives} (predictions with no corresponding ground truth within the threshold), and $FN_{\text{(idf)}}$ indicates \textbf{missed targets} (ground-truth objects not matched by any prediction).
                    
                    For predictions deemed correct under the $1\,\text{m}$ criterion, we further evaluate the localization accuracy by computing the mean Euclidean distance between predicted and ground-truth centers:
                    \begin{equation}
                    \text{LocErr} = \frac{1}{\text{TP}_{\text{(idf)}}} \sum_{i=1}^{\text{TP}_{\text{(idf)}}} \lVert \hat{p}_i - p_i \rVert_2 .
                    \label{eq:loc_error}
                    \end{equation}

                    The results of $\text{Pre}_{\text{(idf)}}$, $\text{Rec}_{\text{(idf)}}$, $\text{F1}_{\text{(idf)}}$, and the mean localization error $\text{LocErr}$ are summarized in Table~\ref{table:localization}.

                    As shown in Table~\ref{table:localization}, our full pipeline achieves promising results in both 3D identification and precise localization, with average F1-scores of 0.848 and 0.839 on the Wuhan and Shanghai datasets, respectively. The mean localization error (LocErr) for correctly localized objects is remarkably low, averaging 0.118\,m and 0.137\,m. This decimeter-level accuracy validates that our geometry-guided optimization strategy effectively ensures high-precision localization, robustly handling the bounding box regression errors and partial occlusions common in 2D detection.

                \begin{table*}[!h]
                    \centering
                    \renewcommand{\arraystretch}{1.0}
                    \caption{3D geo-localization performance across different categories for the Wuhan dataset and Shanghai dataset. $\text{Pre}_{\text{(idf)}}$, $\text{Rec}_{\text{(idf)}}$, and $\text{F1}_{\text{(idf)}}$ are computed with a $1\,\text{m}$ tolerance threshold, and $\text{LocErr}$ denotes the mean Euclidean distance (m) over correctly localized objects.}
                    \label{table:localization}
                    \fontsize{8}{9}\selectfont
                    \begin{tabular}{lcccccccc}
                        \toprule
                        \multirow{2}{*}{Class} & \multicolumn{4}{c}{Wuhan} & \multicolumn{4}{c}{Shanghai} \\
                        \cmidrule(lr){2-5} \cmidrule(lr){6-9}
                        & $\text{Pre}_{\text{(idf)}}$ & $\text{Rec}_{\text{(idf)}}$ & $\text{F1}_{\text{(idf)}}$ & $\text{LocErr}$ (m) & $\text{Pre}_{\text{(idf)}}$ & $\text{Rec}_{\text{(idf)}}$ & $\text{F1}_{\text{(idf)}}$ & $\text{LocErr}$ (m) \\
                        \midrule 
                        Traffic Sign        & 0.770 & 0.891 & 0.826 & 0.106 & 0.895 & 0.874 & 0.884 & 0.139 \\
                        Street Light        & 0.989 & 0.928 & 0.957 & 0.247 & 0.913 & 0.919 & 0.916 & 0.241 \\
                        Signal Light        & 0.909 & 0.991 & 0.948 & 0.100 & 0.884 & 0.955 & 0.919 & 0.134 \\
                        Surveillance Camera & 0.941 & 0.889 & 0.914 & 0.090 & 0.733 & 0.846 & 0.786 & 0.143 \\
                        Barrier Bollard     & 0.852 & 0.831 & 0.841 & 0.064 & --    & --    & --    & --    \\
                        Safety Bollard      & --    & --    & --    & --    & 0.930 & 0.869 & 0.898 & 0.092 \\
                        Fire Hydrant        & 0.800 & 1.000 & 0.889 & 0.086 & 0.889 & 0.842 & 0.865 & 0.110 \\
                        Trash Bin           & 1.000 & 1.000 & 1.000 & 0.069 & 0.929 & 0.975 & 0.951 & 0.111 \\
                        Manhole             & 0.879 & 0.560 & 0.684 & 0.112 & 0.629 & 0.677 & 0.652 & 0.151 \\
                        Traffic Cone        & 0.697 & 0.656 & 0.676 & 0.196 & 0.842 & 0.457 & 0.593 & 0.125 \\
                        Spherical Bollard   & 0.813 & 0.684 & 0.743 & 0.113 & 0.952 & 0.909 & 0.930 & 0.120 \\
                        \midrule
                        Average             & 0.865 & 0.843 & 0.848 & 0.118 & 0.860 & 0.832 & 0.839 & 0.137 \\
                        \bottomrule
                    \end{tabular}
                \end{table*}

                    Analyzing specific categories, large elevated structures (\textit{Street Lights}, \textit{Signal Lights}) achieve high performance (F1 $>$ 0.91) due to clear visibility. Even smaller objects like \textit{Spherical Bollards} perform well. 
                    However, recall drops for \textit{Manholes} and \textit{Traffic Cones}. 
                    \textit{Manholes} suffer from their flat geometry, often appearing in only 1--2 frames with low viewing angles, leading to missed detections and insufficient baselines for triangulation. 
                    \textit{Traffic Cones} are challenged by their linear arrangement along the driving direction. This alignment causes severe mutual occlusion and minimal parallax change, making matching in dense clusters extremely difficult. 
                    Despite these challenges, the consistently low localization error ($<$ 0.2\,m) across all categories confirms the geometric precision of our solver whenever valid associations are established.
            
            \begin{figure*}[!h]
            \centering{
            \includegraphics[width=0.95\textwidth]{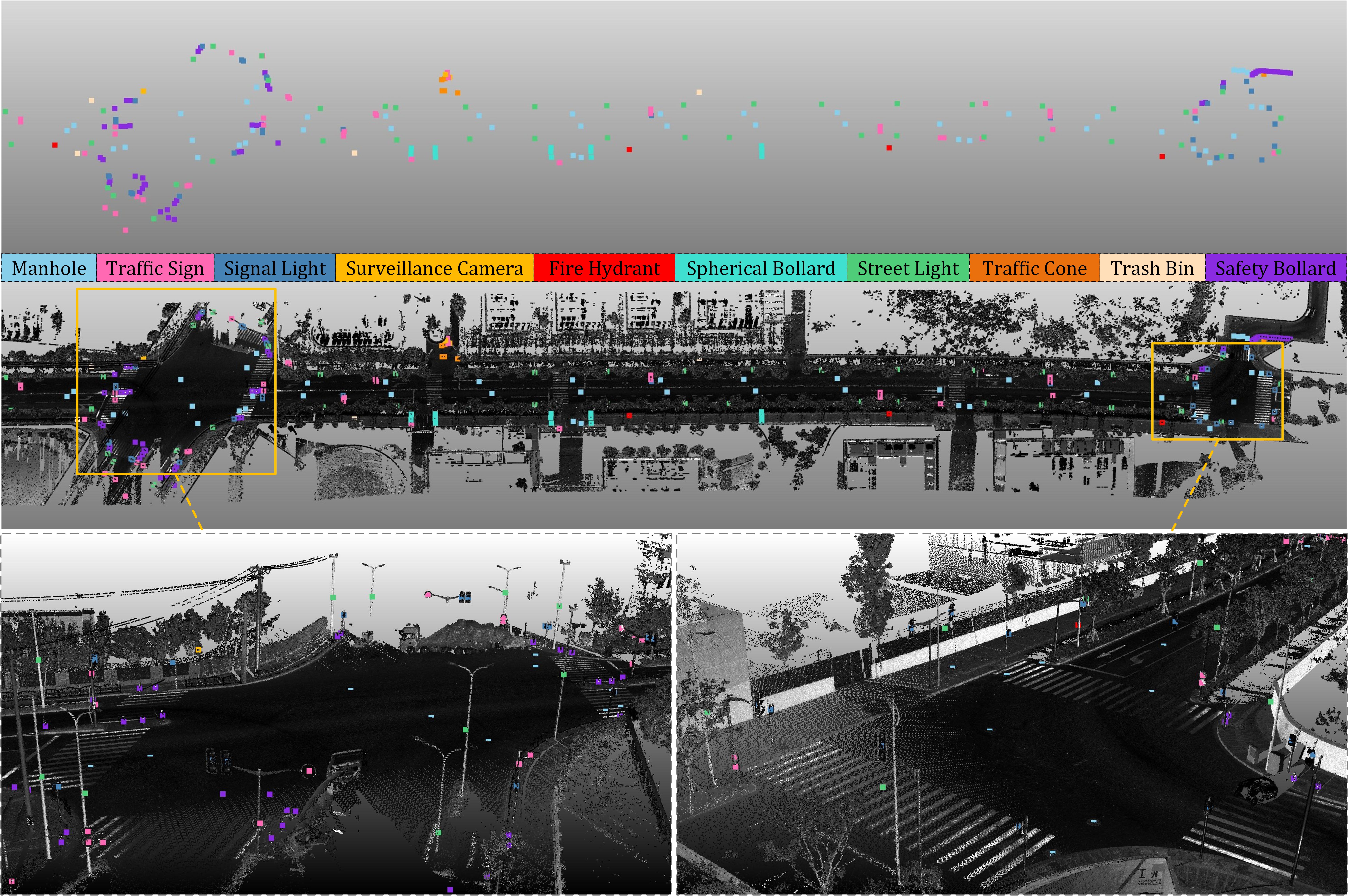}}%
            \caption{Visualization of extracted infrastructure centers in the Shanghai dataset. 
                    The top panel shows the estimated 3D centers, with different colors indicating different object categories. 
                    The middle panel presents the overlay of these centers with the raw LiDAR point cloud (colored by intensity), 
                    providing a global view of the spatial distribution of roadside infrastructure.
                    The bottom images further illustrate a zoomed-in perspective of the intersections.}
            \label {fig_global_shanghai}
            \end{figure*}

            \subsection{Qualitative evaluation}
            To qualitatively assess the effectiveness of our framework, we provide several visualization results.

            Figure~\ref{fig_image_detect} presents the qualitative results of the detection module on the test set. It can be seen that the fine-tuned model adapts well to the panoramic field of view, accurately detecting multiple types of infrastructure in typical urban street scenes. Notably, the method demonstrates adaptability to objects of varying scales, successfully covering large nearby street lights while accurately recalling smaller, distant targets such as bollards. These high-quality 2D detection results establish the foundation for subsequent 3D localization.
            
            Next, we present more detailed visualizations in Figure~\ref{fig_detail_categories}, where we show the raw point clouds and the extracted centers for selected categories. The results confirm that the estimated centers possess accurate 3D spatial coordinates, consistent with the underlying point cloud structures.
            
            Furthermore, Fig.~\ref{fig_global_shanghai} provides a global view of the Shanghai dataset, where the extracted centers of various categories are shown in different colors and overlaid on the raw LiDAR point cloud. This comprehensive visualization provides a clear impression of the overall localization results across the entire scene.

\begin{figure*}[!h]
            \centering{
            \includegraphics[width=0.95\textwidth]{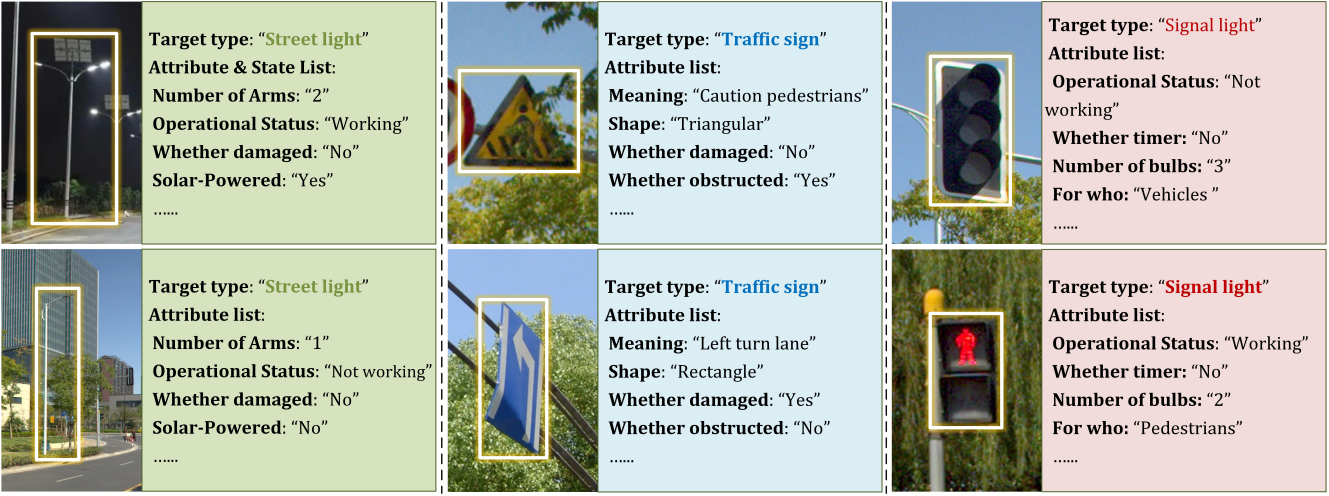}}%
            \caption{Examples of fine-grained attribute and operational state recognition by the proposed VLM agent. For each infrastructure target (e.g., street light, traffic sign, signal light), the agent infers a structured attribute list including static properties (e.g., shape, number of arms) and dynamic states (e.g., operational status, damage condition).}
            \label {fig_vlm_examples}
            \end{figure*}

            Finally, our state-discriminative VLM agent enables comprehensive semantic understanding of infrastructure assets. As illustrated in Figure~\ref{fig_vlm_examples}, for each detected object, the agent outputs a structured list of attributes and operational states. For instance, it can identify not only that a street light is “Working” but also its “Number of Arms”, whether it is “Solar-Powered”, or if it is “Damaged”; for a traffic sign, it can discern its “Meaning”, “Shape”, and whether it is “Obstructed”; and for a signal light, it can determine its “Operational Status”, “Number of bulbs”, and “Target user group” (e.g., “Vehicles” or “Pedestrians”). This level of detail provides critical context for maintenance prioritization, anomaly detection, and cost estimation, directly supporting intelligent urban management and budget planning.

            \subsection{Comparison and ablation studies}
                We conduct comprehensive experiments to evaluate the effectiveness of our proposed framework. While some prior works address similar tasks, they are either closed-source or not tailored for the sparse, wide-baseline setting of our dataset, making direct reproduction and fair comparison infeasible. Consequently, we establish two strong baselines representing standard approaches to validate our contributions:
                (1) Zero-shot Foundation Model: The original Grounding DINO model without domain adaptation, serving as a baseline for open-set detection.
                (2) Local-Association Chaining: A standard matching strategy that relies solely on local pairwise predictions to link observations across frames transitively, without global geometric optimization or refinement. This represents the conventional approach for extending pairwise matching to multi-view sequences.
                By comparing against these baselines, we simultaneously perform ablation studies to validate the contributions of our key components: LoRA fine-tuning and Geometry-Guided Matching Refinement.

                    \begin{table}[!h]
                        \centering
                        \renewcommand{\arraystretch}{1.0}
                        \caption{Quantitative comparison between the \textbf{Zero-shot Baseline} (original Grounding DINO) and \textbf{SVII-3D} (equipped with LoRA Fine-tuned Model). Metrics include 2D detection performance (AP@0.5, Pre$_{\text{(det)}}$, Rec$_{\text{(det)}}$) and the impact on downstream \textbf{3D discovery and localization} (Pre$_{\text{(idf)}}$, Rec$_{\text{(idf)}}$, LocErr).}
                        \label{table_ablation_detection}
                        \fontsize{8}{9}\selectfont
                        \resizebox{0.50\textwidth}{!}{%
                        \begin{tabular}{lccccccc}
                        \toprule
                        \multirow{2}{*}{Dataset} & \multirow{2}{*}{Method} & \multicolumn{3}{c}{2D Detection} & \multicolumn{3}{c}{3D Identification and Localization} \\
                        \cmidrule(lr){3-5} \cmidrule(lr){6-8}
                        & & AP@0.5 & Pre$_{\text{(det)}}$ & Rec$_{\text{(det)}}$ & Pre$_{\text{(idf)}}$ & Rec$_{\text{(idf)}}$ & LocErr \\
                        \midrule
                        \multirow{2}{*}{Wuhan} 
                        & Baseline & 25.4 & 42.7 & 45.0 & 0.638 & 0.523 & 0.225 \\
                        & \textbf{SVII-3D}  & \textbf{73.1} & \textbf{74.5} & \textbf{80.6} & \textbf{0.865} & \textbf{0.843} & \textbf{0.118} \\
                        \midrule
                        \multirow{2}{*}{Shanghai} 
                        & Baseline & 26.5 & 44.9 & 45.3 & 0.583 & 0.633 & 0.234 \\
                        & \textbf{SVII-3D}  & \textbf{77.5} & \textbf{80.2} & \textbf{84.2} & \textbf{0.860} & \textbf{0.832} & \textbf{0.137} \\
                        \bottomrule
                        \end{tabular}%
                        }
                    \end{table}
                
                \subsubsection{Comparison with Zero-shot Foundation Model}
                    We first validate the necessity of domain adaptation by comparing our LoRA fine-tuned detector against a Zero-shot Baseline (the original pre-trained Grounding DINO model without fine-tuning). The quantitative results are detailed in Table~\ref{table_ablation_detection}.
                    
                    \textbf{Improvement in 2D Detection.} The baseline model, despite its strong open-set capabilities, struggles with domain-specific infrastructure categories in complex street scenes, achieving an average AP@0.5 of only 25.4\% on Wuhan and 26.5\% on Shanghai. In contrast, our SVII-3D model yields a dramatic performance boost, tripling the AP@0.5 to 73.1\% and 77.5\%, respectively. This confirms that lightweight fine-tuning is essential for bridging the domain gap between general pre-training data and specific urban maintenance targets.
                    
                    \textbf{Cascading Effect on 3D Localization.} The improvement in 2D perception directly translates into superior downstream 3D localization. As shown in Table~\ref{table_ablation_detection}, the high-quality bounding boxes provided by our model enable more reliable ray triangulation. Compared to the baseline, our approach improves 3D Identification Precision by over 20\% (from $\sim$0.6 to $\sim$0.86) and nearly halves the mean Localization Error (LocErr) from $\sim$0.23\,m to $\sim$0.12\,m. This demonstrates that accurate 2D detection is the prerequisite foundation for high-precision 3D mapping.

                    \begin{table}[!h]
                        \centering
                        \renewcommand{\arraystretch}{1.0}
                        \caption{Quantitative comparison between the \textbf{Local-Association Chaining baseline} and \textbf{SVII-3D} (incorporating Geometry-Guided Refinement). Metrics: Homogeneity (Homo.) and Completeness (Comp.) evaluate clustering quality; Identification Precision (Pre$_{\text{(idf)}}$) and Recall (Rec$_{\text{(idf)}}$) assess \textbf{3D identification performance}, while Error (LocErr) measures \textbf{localization accuracy}.}
                        \label{table_ablation_refinement}
                        \fontsize{8}{9}\selectfont
                        \resizebox{0.50\textwidth}{!}{%
                        \begin{tabular}{lcccccc}
                        \toprule
                        \multirow{2}{*}{Dataset} & \multirow{2}{*}{Method} & \multicolumn{2}{c}{Clustering} & \multicolumn{3}{c}{3D Identification and Localization} \\
                        \cmidrule(lr){3-4} \cmidrule(lr){5-7}
                        & & Homo. & Comp. & Pre$_{\text{(idf)}}$ & Rec$_{\text{(idf)}}$ & LocErr \\
                        \midrule
                        \multirow{2}{*}{Wuhan} 
                        & Baseline & 0.936 & 0.886 & 0.562 & 0.734 & 0.144 \\ 
                        & \textbf{SVII-3D} & \textbf{0.993} & \textbf{0.975} & \textbf{0.865} & \textbf{0.843} & \textbf{0.118} \\
                        \midrule
                        \multirow{2}{*}{Shanghai} 
                        & Baseline & 0.937 & 0.825 & 0.456 & 0.751 & 0.160 \\ 
                        & \textbf{SVII-3D} & \textbf{0.989} & \textbf{0.975} & \textbf{0.860} & \textbf{0.832} & \textbf{0.137} \\
                        \bottomrule
                        \end{tabular}
                        }
                    \end{table}

                \subsubsection{Comparison with Local-Association Chaining}
                    We evaluate the impact of our Geometry-Guided Matching Refinement by comparing it against a Local-Association Chaining baseline (denoted as ``Baseline'' in Table~\ref{table_ablation_refinement}). This baseline relies solely on transitivity to link pairwise predictions into clusters without global geometric constraints.
                    
                    \textbf{Improvement in Observation Grouping.} The baseline approach is prone to structural errors due to error propagation in local associations. As shown in Table~\ref{table_ablation_refinement}, applying the refinement strategy of SVII-3D yields consistent gains in clustering quality: Homogeneity increases by approximately 5.7\% on Wuhan and 5.2\% on Shanghai, while Completeness improves significantly by 8.9\% and 15.0\%, respectively. This confirms that our strategy effectively acts as a global corrector, resolving over-matching by splitting geometrically inconsistent rays and fixing under-matching by merging fragmented observations.
                    
                    \textbf{Impact on 3D Localization.} The refinement step proves to be the decisive factor for geo-localization accuracy.
                    In the baseline setting, the lack of geometric verification often leads to the merging of distinct physical objects into a single cluster. This results in a geometrically shifted center (e.g., the midpoint between two objects) that exceeds the 1\,m tolerance, causing a severe drop in Identification Precision (e.g., only 0.456 on Shanghai).
                    By enforcing strict geometric consistency, our algorithm successfully prunes these outliers and rectifies the cluster structures. Consequently, Identification Precision surges by over 30\% on Wuhan ($0.562 \rightarrow 0.865$) and 40\% on Shanghai ($0.456 \rightarrow 0.860$). Furthermore, Identification Recall sees a substantial boost ($\sim$10\%) as valid objects are recovered from previously discarded singletons through the merging process.

        \section{Conclusions}
            This study presents SVII-3D, a comprehensive framework for automated roadside infrastructure inventory using low-cost, sparse street-view imagery. By synergizing LoRA fine-tuned open-set detection, spatial-attention-based cross-view matching, and a novel geometry-guided refinement strategy, our approach effectively overcomes the challenges posed by wide baselines and visual ambiguities in complex urban environments.
            
            Quantitative experiments on city-scale datasets from Wuhan and Shanghai demonstrate that the proposed pipeline achieves robust performance. The geometry-guided global refinement mechanism significantly reduces structural matching errors, enabling decimeter-level 3D localization accuracy (with a mean error of approximately 0.12\,m) and high retrieval completeness. Furthermore, the rigorous cross-city evaluation protocol confirms the method's strong generalization capability, verifying that the model learns intrinsic geometric invariance rather than overfitting to specific city layouts.

            Beyond geometric precision, the integration of our State-discriminative VLM Agent marks a shift from static mapping to semantic comprehension. By leveraging RAG and expert knowledge injection, the system extracts fine-grained attributes and diagnoses operational states (e.g., distinguishing between structural damage and surface dirt), providing actionable insights for infrastructure maintenance and facility management. Collectively, this framework offers a scalable solution for building high-fidelity semantic 3D digital twins, serving as a critical technology for automating the survey and inspection of urban built environments.
            
            Future work will introduce a VLM-driven ``Digital Quality Inspector'' to enable machine self-verification. By fine-tuning on synthesized error samples, this autonomous agent will detect perception failures (e.g., false positives) by verifying semantic consistency between raw data and model outputs, thereby achieving a trustworthy, closed-loop automated inventory system.
            
        \section *{Declaration of Interests}
            The authors declare that they have no known competing financial interests or personal relationships that could have appeared to influence the work reported in this paper.
        
        \section *{Acknowledgment}
            This work was supported by the National Natural Science Foundation of China (No. 42571521).
        
\bibliography{rmfile}

\end{document}